\documentclass[a4paper]{article}
\usepackage[margin=1in]{geometry} 

\usepackage{amsmath}
\usepackage{amsthm}
\usepackage{amssymb}
\newcommand{\Tau}{\mathrm{T}}
\usepackage[utf8]{inputenc}
\usepackage{hyperref}
\hypersetup{
	unicode,
	pdfauthor={Author One, Author Two, Author Three},
	pdftitle={A simple article template},
	pdfsubject={A simple article template},
	pdfkeywords={article, template, simple},
	pdfproducer={LaTeX},
	pdfcreator={pdflatex}
}

\usepackage[sort&compress,numbers,square]{natbib}
\theoremstyle{plain}

\theoremstyle{definition}

\usepackage{graphicx, color}
\graphicspath{{fig/}}

\usepackage{algorithm, algpseudocode} 
\usepackage{mathrsfs} 

\usepackage{lipsum}

\title{Learnable Data Augmentation for One-Shot Unsupervised Domain Adaptation}
\author{Julio Ivan Davila Carrazco$^{1,4}$ 
\and Pietro Morerio$^{1}$
\and Alessio Del Bue$^{1}$
\and Vittorio Murino$^{1,2,3}$}

\date{
$^1$Pattern Analysis and Computer Vision (PAVIS), Italian Institute of Technology, Genoa, Italy \texttt{\{julio.davila, pietro.morerio, alessio.delbue, vittorio.murino\}@iit.it}\\%
$^2$Department of Computer Science, University of Verona, Verona, Italy\\%
$^3$Department of Computer Science and Technology, Bioengineering, Robotics and Systems Engineering, University of Genova, Genoa, Italy\\%
$^4$Department of Marine, Electrical, Electronic and Telecommunications Engineering Robotics and Systems Engineering, University of Genova, Genoa, Italy\\%
}

\begin{document}
	\maketitle
	
\begin{abstract}
This paper presents a classification framework based on learnable data augmentation to tackle the One-Shot Unsupervised Domain Adaptation (OS-UDA) problem. OS-UDA is the most challenging setting in Domain Adaptation, as only one single unlabeled target sample is assumed to be available for model adaptation. Driven by such single sample, our method LearnAug-UDA learns how to augment source data, making it perceptually similar to the target. As a result, a classifier trained on such augmented data will generalize well for the target domain. To achieve this, we designed an encoder-decoder architecture that exploits a perceptual loss and style transfer strategies to augment the source data. Our method achieves state-of-the-art performance on two well-known Domain Adaptation benchmarks, DomainNet and VisDA. The project code is available at  \href{https://github.com/IIT-PAVIS/LearnAug-UDA}{https://github.com/IIT-PAVIS/LearnAug-UDA}

\end{abstract}

\section{Introduction}
\label{sec:intro}

Although deep learning empowers us to tackle intricate tasks with exceptional performance, it comes with an inherent trade-off: a voracious appetite for copious amounts of data for effective model training. To overcome this, new areas of research were born, specifically designed to cope with either the scarcity of labeled data, the availability of unlabeled data only, or situations where only few samples can be exploited. In these settings, Domain Adaptation focuses on the task of transferring knowledge from a (usually) richly annotated Source domain to a Target domain, where data is assumed to be scarce or labelling is absent, in any case 
insufficient to train a model from scratch. When target data is completely unlabeled, we deal with the research area of Unsupervised Domain Adaptation (UDA). 
In UDA, the scarcity of target samples could be a further issue. Few-shot Unsupervised Domain Adaptation (FS-UDA) focuses on solving the DA problem where only few labeled source samples are available \cite{yue2021prototypical,yang2022few}, while one-shot Unsupervised Domain Adaptation (OS-UDA) is the case where target data is reduced to \textit{only one unlabeled sample} \cite{luo2020adversarial, carrazco2023targetdriven}. Last, an even more extreme setting is the one referred to as Domain Generalization \cite{muandet2013domain,li2018domain, Carlucci_2019_CVPR}, where no data whatsoever is available for the target domain(s), yet one would like the model to generalize as best as it can to unseen domains.

To the best of our knowledge, there are only two methods focusing on solving OS-UDA. Adversarial Style Mining (ASM) \cite{luo2020adversarial} leverages style transfer to synthesize target samples to be used for training a classifier.
Target-driven One-shot Unsupervised Domain Adaptation (TOS-UDA)\cite{ carrazco2023targetdriven} applies a two-step strategy to learn a set of fixed transformations guided by a perceptual loss. 

In this paper, we propose a new approach to tackle the OS-UDA task. We design an original method to \textit{learn} augmentations for the source samples to resemble the (only) target data by exploiting style transfer strategies. To ensure the quality and fidelity of the learned augmentations, we employ a perceptual loss function. By doing so, we guide the learned augmentations to closely align with the target domain. 
In summary, in our work:.
\begin{itemize}
    \itemsep0em
    \item We present LearnAug-UDA a novel approach for tackling One-shot Unsupervised Domain Adaptation setting, which is a challenging UDA task where only one single unlabeled target sample is available. 
    \item The proposed architecture is composed by two encoder-decoder modules to learn augmentations by enforcing a perceptual similarity and conditioning with two types of methods: Mixup\cite{zhang2017mixup} and Style disentanglement\cite{Zhang_2018_CVPR, huang2017arbitrary}. 
    \item We achieve state-of-the-art results for OS-UDA on DomainNet and VisDA benchmarks. 
\end{itemize}

The rest of the paper is organized as follows. In Section \ref{sec:related_work}, we illustrate the related work, in comparison with our proposed solution. In Section \ref{sec:methodology}, we detail the description of our proposed model and its component modules. In Section \ref{sec:experiments}, we introduce the domain adaptation benchmarks used for testing, and we present the results obtained by our method. In Section \ref{sec:conclusions}, we draw the conclusions of the work, while commenting our proposed approach in relation to both the methodological choices and the experimental analysis.

\section{Related work}
\label{sec:related_work}
\noindent
\textbf{One-Shot Unsupervised Domain Adaptation (OS-UDA)}: as mentioned above, OS-UDA refers to the scenario in Unsupervised Domain Adaptation where the available samples of the Target domain are reduced to only one unlabeled sample and the literature focusing on this setting is very limited. In \cite{luo2020adversarial}, They introduced \textit{Adversarial Style Mining} (ASM). ASM leverages style transfer to augment source samples with the style of a single target sample, by introducing a Random Adaptive Instance Normalization (RAIN) module, which in practice performs style transfer. However, RAIN is pretrained for such a task, with extra information from an external dataset (wikiArts). Therefore, while a single target sample is used to drive RAIN at inference time, extra data is actually exploited to learn the style transfer task. ASM's training pipeline is not end-to-end and the additional dataset for pre-training needs to be manually selected based on task and characteristics of source and target dataset. In \cite{carrazco2023targetdriven}, They introduced \textit{Target-driven One-shot Unsupervised Domain Adaptation} (TOS-UDA) which applies adversarial training and a perceptual loss to guide an augmentation module to learn the parameters of two fixed transformations. TOS-UDA augments source samples to train a classifier, which is expected to generalize well for target data. 

An alternative to using domain adaptation methods is to re-purpose methods that does not tackle the domain adaptation problem per se. \textit{TeachAugment} \cite{Suzuki_2022_CVPR} is presented as a method for data augmentation to improve model generalization. TeachAugment augments data samples by performing two types of fixed transformations: a geometrical and color transformations. The parameters of such transformations are then defined by an optimization-based approach and an adversarial training strategy is used to drive the  process. This strategy aims to generate augmented samples that are ``hard'' for the model to learn from. Our approach, although drawing similarities from the aforementioned methods, obtains better results by exploiting the unused potential that the other methods omitted. First of all, we applied an encoder-decoder architecture for our augmentation module. This allows the model to learn better augmentations as it is not bound to a set of fixed transformations. Furthermore, we introduce a modified perceptual loss, able to measure the similarity in both style and content.

\textbf{Style transfer}: Style transfer refers to task of synthesizing an image that resembles the characteristics in style from an image and the content from a different image. In \cite{gatys2015neural}, a first approach to use Convolutional Neural Networks (CNNs) for style transfer was presented. This process was later referred as Neural Style Transfer (NST). NST disentangles content and style by using features representations of different layers of a VGG network \cite{simonyan2014very}. In \cite{johnson2016perceptual}, they proposed a perceptual loss for training a feed-forward network for the style transfer task. The perceptual loss measures the perceptual similarity between the synthesized image, content image, and the style image. The first methods solving style transfer were only capable of transfer one specific style \cite{gatys2015neural, johnson2016perceptual}. Later, methods \cite{dumoulin2016learned,Zhang_2018_ECCV_Workshops} raised the expectations by learning different styles from a set of target images. Finally, the introduction of arbitrary style transfer allowed to perform style transfer without being bound to a set of style targets \cite{Zhang_2018_CVPR, huang2017arbitrary, Liu_2021_ICCV}. We can regard our proposed augmentation method as a style transfer module, which has to deal with the difficult task of learning style from a single target image and transferring it to the source samples.

\section{Methodology}
\label{sec:methodology}

\begin{figure}[ht]
\centering
\includegraphics[width=\textwidth]{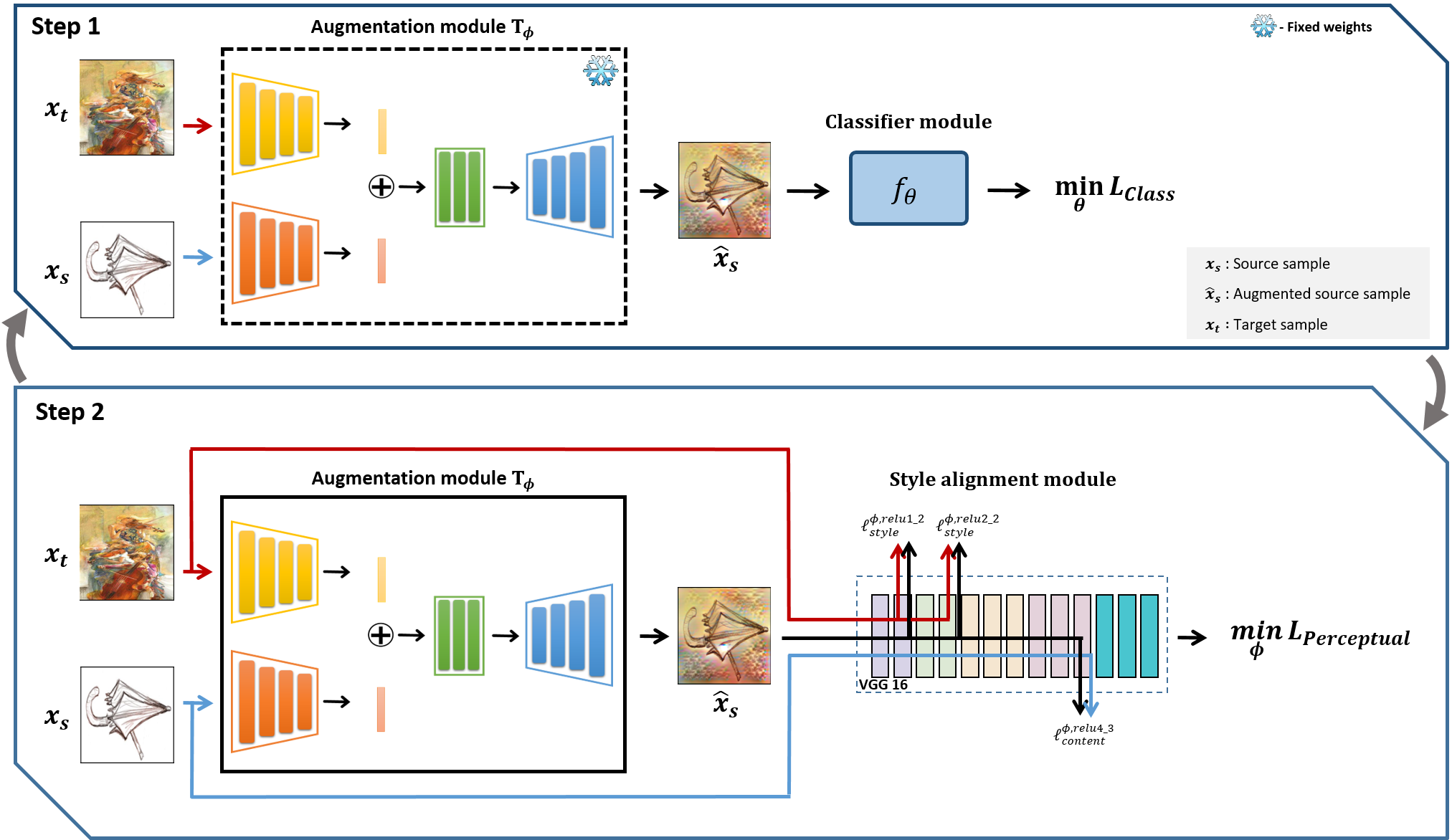}
\caption{Our proposed approach with its three elements: Augmentation module (AUM), the Style Alignment Module (SAM) and the Classifier Module (CM). The training strategy focuses on alternating between updating the CM and AUM. In step 1, the CM is trained by minimizing a cross-entropy loss. In step 2, AUM learns to synthesize augmentations guided by SAM whom measures the perceptual similarity with the Target sample.}
\label{fig:architecture}
\end{figure}

Our proposed model is constituted by three elements: an augmentation module, a style alignment module and a classifier. Each of these elements focuses on a specific task. The Augmentation Module (AUM) augments source samples to resemble the Target domain. The AUM achieves this by conditioning the image synthesis on the only target sample that is available. The Style Alignment Module (SAM) is responsible of measuring the perceptual similarities (at different levels) between Source samples, the Target sample, and the synthesized augmented samples. Similar to style transfer, AUM and SAM work together to learn the augmentations for the Source data (see Fig. \ref{fig:architecture}, Step 2). In other words, SAM guides the learning process of AUM to obtain augmentations that have a similar style as the Target domain. Our intuition is that the classifier trained on these augmented samples will generalize better for the Target domain (see Fig. \ref{fig:architecture}, Step 1). A two-step training strategy is adopted (see Fig. \ref{fig:architecture}). In the first step, the AUM synthesizes augmented samples for the classifier to train on. In this step, only the classifier is updated via back propagation while the AUM's weights remain frozen. In the second step, AUM is trained by learning augmentations that favour a similarity with the Target domain's style. More detalis are provided in the following.

\subsection{Augmentation module}
The Augmentation Module (AUM) manages the synthesis of augmented samples used for training the Classifier module (CM). We designed two variations of the augmentation module architecture to explore different approaches for achieving style transfer in data augmentation.

\begin{figure}[h]
\centering
\includegraphics[width=\textwidth]{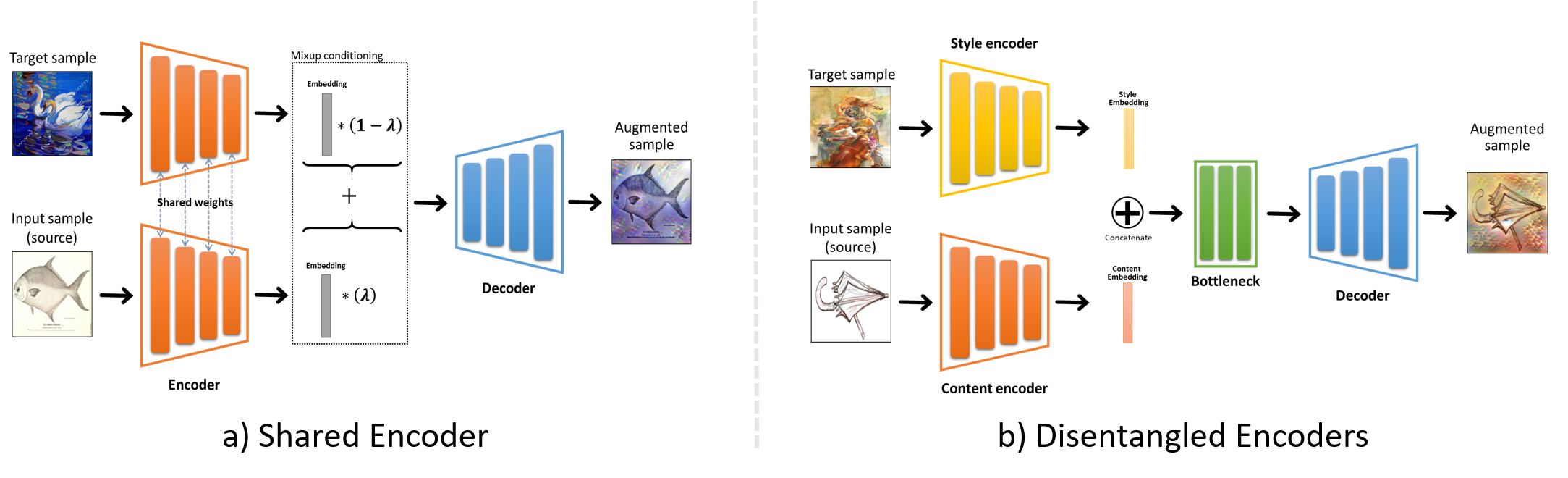}
\caption{The Augmentation Module (AUM) in its two variations: Shared encoder (a) and Disentangled encoders (b). The Shared encoder applies the same encoding to both the source and target samples, while performing style conditioning through mixup at the bottleneck. The Disentangled encoders treat style and content as separate elements within the image, with each individual encoder learning a distinct feature representation.}
\label{fig:AugmentationModule}
\end{figure}

The first variation, the Shared Encoder (see Fig. \ref{fig:AugmentationModule}a), is a simple encoder-decoder architecture. Both style and content samples are encoded using the same encoder, with conditioning performed at the bottleneck through Mixup between the embeddings \cite{zhang2017mixup, huang2017arbitrary}. Namely a convex combination of source and target sample is fed as input to the decoder, in order to synthesize the augmented sample. We designed the Share Encoder to exploit a Mixup-inspired data augmentation approach where we wanted the resulting feature maps to contain characteristics of both domains.

The second variation consists of Disentangled Encoders (see Fig. \ref{fig:AugmentationModule}b).  This design leverages the concept of domain separation \cite{bousmalis2016domain}, the intuition being that different domains exhibit shared characteristics and domain-specific characteristics. Similarly, image style and content can be considered separate aspects. Therefore, our disentangled encoders consist of a Style encoder and a Content encoder. In this case, the intuition is that differences in style are responsible for the performance drop of the source classifier tested on target data. Each encoder specializes in learning different aspects of the image and generates an embedding representing that particular aspect. To synthesize an augmented sample, the concatenated embeddings are passed to the Bottleneck module, to reduce dimensionality, and subsequently a decoder generates the augmented sample. In contrast to the shared encoder, we do not perform Mixup between embeddings with this architecture, as fusion is left to the bottleneck module. To improve the disentanglement between style and content, we introduce a reconstruction loss which allows the encoders to see samples of both domains. The reconstruction loss is defined as:
\begin{equation}
  L_{rec} = \lVert \Tau_\phi(x,x) -  x \rVert^2_2
  \label{eq:reconstruction_loss}
\end{equation}
\noindent
where $\Tau_\phi( \cdot)$ is the augmentation module, $x$ is an input sample from the source or target data.

\subsection{Style alignment module}
The Style alignment module (SAM) is in charge of enforcing a perceptual similarity in style and content between a source sample, the only target sample and the synthesized augmented sample. SAM imposes a similarity in style with respect to the target sample. By doing so, AUM learns to synthesize augmented samples for the Target domain. Furthermore, by imposing a similarity in content with respect to the input source sample, i.e. the shared characteristics between source and target, the classifier can use labels from the input source to train for the target domain. SAM consists of a frozen VGG-16 \cite{simonyan2014very} from which feature maps of selected layers are extracted to evaluate the perceptual similarity \cite{johnson2016perceptual}. The perceptual loss comprises two type of losses: a style and content loss. The style loss evaluates similarity between aspects such as color, texture, and common patterns. Gatys et al. \cite{gatys2016image} defined the representation of style as the correlation between filter responses of the feature map. These feature correlations are represented as a Gram matrix which is defined as:
\begin{equation}
  G_{j}(x)=\dfrac{1}{C_j H_j W_j}\sum^{H_j}_{h=1}\sum^{W_j}_{w=1}\mathscr{H}_j(x) \mathscr{H}^{T}_{j}(x)
  \label{eq:gram_matrix}
\end{equation}
\noindent
where $x$ is the input image, $\mathscr{H}_j$ represents the feature map from the $j$-layer of the SAM; $C_j$, $H_j$, and $W_j$ denote the number of channels, the height, and the width of $\mathscr{H}_j$, respectively. The Gram matrix is calculated for the target and the augmented samples. The style loss of a single layer is calculated as:
\begin{equation}
  \ell^{style}_i(\hat{x_s}, x_t) = \lVert G_i(\hat{x_s}) -  G_i(x_t) \rVert^2_F \hspace{0.5em} \textrm{, where} \hspace{0.5em} \hat{x_s} = \Tau_\phi(x_s,x_t)
  \label{eq:style_layer_loss}
\end{equation}
\noindent
where $\Tau_\phi(\cdot)$ is the Augmentation module, $\hat{x_s}$ is then the augmented sample, $x_s$ represents a source sample, $x_t$ indicates the target sample, $\lVert \cdot \rVert^2_F$ is the Frobenius norm, and $i$ represents the selected layers of SAM. For the content loss which focuses on preserving the semantic content, i.e. the actual objects present in the scene, we define it as:
\begin{equation}
 \ell^{content}_j(\hat{x_s}, x_s) = \dfrac{1}{C_j H_j W_j}\lVert \mathscr{H}_j(\hat{x_s}) -  \mathscr{H}_j(x_s) \rVert^2_2
  \label{eq:content_layer_loss}
\end{equation}
\noindent
where $\mathscr{H}_j$ represents the feature map from the $j$-layer of the SAM; $C_j$, $H_j$, and $W_j$ denote the number of channels, the height, and the width of $\mathscr{H}_j$, respectively. As established by \cite{johnson2016perceptual, gatys2016image}, multiple layers could extract the information regarding the style and content. So, it stands to reason to add up the individual loss of different layers to better capture the style and content information. Therefore, these cumulative losses are defined as:
\begin{alignat}{2}
  &L_{style} = \sum_{i}\ell^{style}_i(\hat{x_s},x_t), 
  &\qquad
  &L_{content} = \sum_{j}\ell^{content}_j(\hat{x_s},x_s)
\end{alignat}
\noindent
where $i$ and $j$ represent the selected layers of SAM for evaluating style and content, respectively. Finally, The perceptual loss is calculated as:
\begin{equation}
  L_{perceptual\_G} = L_{style} + L_{content}
  \label{eq:perceptual_loss}
\end{equation}
The perceptual loss, as defined in the previous equations, has limitations when performing style transfer with only one target sample. Naively enforcing a perceptual similarity against the single
target sample leads to locally injecting its feature’s patterns
always in the same position. The synthesized samples tend to exhibit characteristics related to the target sample's content, such as shapes and contours. 
To address this issue, we propose a new variation of the perceptual loss that incorporates an average pooling operation over the feature maps. By using a spatial average
pooling operation, we instead enforce a global perceptual
similarity of the feature maps. This operation introduces a smoothing effect on the reduced feature map, leading to a smoother synthesized sample. We apply this new variation to both the style and content losses, defining the new loss as follows:
\begin{equation}
  \ell_i(\hat{x_s}, x_c) = \dfrac{1}{C H W}\lVert AvP(\mathscr{H}_i(\hat{x_s})) -  AvP(\mathscr{H}_i(x_c)) \rVert^2_2
  \label{eq:avp_layer_loss}
\end{equation}
\noindent
where $\hat{x_s}$ represents the augmented sample, $x_c$ represents the sample to compare against with, i.e. the target or input source sample, $\mathscr{H}_i$ represents the feature map from the $i$-layer of the SAM, $AvP$ specifies the average pooling operation,  $C$, $H$, and $W$ denote the number of channels, the height, and the width of reduced feature map after the average pooling operation, respectively. Finally, the new perceptual loss is redefined as:
\begin{equation}
  L_{perceptual\_AvP} = \sum_{i}\ell_i(\hat{x_s},x_s) + \sum_{j}\ell_j(\hat{x_s},x_t)
  \label{eq:perceptual_avp_loss}
\end{equation}
\noindent
where $x_s$ represents input source sample, $x_t$ represents the only target sample, $i$ and $j$ represent the selected layers of SAM for evaluating style and content, respectively.
\subsection{Classifier module}
The Classifier module (CM) is trained on the augmented samples and the labels of their corresponding input source sample. The intuition behind training the classifier with augmented samples is that by doing so it should generalize well for the target domain. The classifier is trained in supervised manner by minimizing a traditional cross entropy loss:
\begin{equation}
  L_{CE} = -\sum_{x_s \in X}y_k\log f_\theta(\Tau_\phi(x_s,x_t)) 
  \label{eq:cross_entropy}
\end{equation}
\noindent
where $X$ is the source sample set, $y$ $\in$ $\{0,1\}^K$ denotes the one-hot ground-truth vector, $K$ is the number of classes, $\Tau_\phi(\cdot)$ denotes the AUM with parameters $\phi$, and $f_\theta(\cdot)$ represents the classifier with parameters $\theta$.
\subsection{Training strategy}
A two-step training strategy alternates the learning of the CM and the AUM to enhance the performance and generalization of the model (see Fig. \ref{fig:architecture}). This strategy is applied to stabilize the learning process. This process will allow the individual modules to update their parameters without causing drastic changes in the other modules. The division of task leads to a more effective and specialized learning in each module. 

In the first step, the classifier is trained on the augmented samples to learn the underlying patterns and features. The CM is trained by minimizing Eq. \ref{eq:cross_entropy}. During step 1, the weights of AUM are frozen, allowing the backpropagation process to update only the Classifier Module (CM).

In the second step, AUM is trained, guided by SAM, to learn augmentations that resemble the Target domain. AUM can be trained using either variation of the perceptual loss (Eq. \ref{eq:perceptual_loss} or Eq. \ref{eq:perceptual_avp_loss}). When the Disentangled encoders are used, the reconstruction loss (eq. \ref{eq:reconstruction_loss}) can be used together with the perceptual loss to improve AUG's performance. The learning process of AUM can be controlled by conducting several iterations solely focused on step 1. 
\section{Experiments}
\label{sec:experiments}
\subsection{Datasets}
To assess the performance of our proposed method, we evaluate it on two well-known benchmarks for domain adaptation: \textbf{DomainNet} \cite{peng2019moment} and \textbf{VisDA} \cite{peng2017visda}. DomainNet consists of four distinct domains: Real, Painting, Clipart, and Sketch. Following the evaluation procedure established in \cite{Saito_2019_ICCV}, we conduct our assessment across seven domain adaptation tasks: Real to Clipart (\textbf{R} $\rightarrow$ \textbf{C}), Real to Painting (\textbf{R} $\rightarrow$ \textbf{P}), Real to Sketch (\textbf{R} $\rightarrow$ \textbf{S}), Painting to Clipart (\textbf{P} $\rightarrow$ \textbf{C}), Painting to Real (\textbf{P} $\rightarrow$ \textbf{R}), Clipart to Sketch (\textbf{C} $\rightarrow$ \textbf{S}), and Sketch to Painting (\textbf{S} $\rightarrow$ \textbf{P}). The VisDA dataset, on the other hand, comprises two domains: Synthetic and Real. The data is divided into three splits: training, validation, and testing. The training split represents the synthetic domain, while both the validation and testing splits consist of real domain samples sourced from different image datasets (COCO \cite{lin2014microsoft} and the Youtube Bounding Box dataset \cite{real2017youtube}). For our experiments, we treat the validation split as the training set for the Real domain, as it is from this set that we obtain our target samples.

\subsection{Setup}
The pipeline training setup is as follows: the classifier CM is a ResNet-101 pretrained on ImageNet. AUM was tested in its two variations: Shared encoder (SE) and Disentangled encoders (DE). SAM, is a VGG-16 \cite{simonyan2014very} pretrained on ImageNet. In the VGG-16, the features maps from the layers $relu1\_2$ and $relu2\_2$ were used for the style loss while the features map from layer $relu4\_3$ was used for the content loss. AUM's layers descriptions will be presented on the supplementary material. All the experiments were run five times. We present mean and std for DomainNet and only the mean for VisDA. VisDA results' standard deviations will be presented on the supplementary.
\subsection{Implementation details}

To train LearnAug-UDA, we employed two optimizers. Each of them focusing on individually updating the CM and AUG modules. The CM is trained using stochastic gradient descend  (SDG) while AUG is trained using the AdamW optimizer. The perceptual loss' weights for layers $relu1\_2$, $relu2\_2$, and $relu4\_3$ were set as 0.25,1.0, and 1.0 respectively. The model was trained for 20 epochs with a learning rate of $1e-4$ for the CM optimizer and $1e-3$ for the AUG optimizer, batch size of 8 samples with input size of $224x224$. When training with SE and Mixup, the parameters for the beta distribution were set as 5.0 for the alpha value and 1.0 for the beta value. Futher information can be found on the supplementary material and the code repository of LearnAug-UDA: \href{https://github.com/IIT-PAVIS/LearnAug-UDA}{https://github.com/IIT-PAVIS/LearnAug-UDA}.

\begin{table}[]
\centering
\caption{\small{Classification accuracies of our proposed method for DomainNet on seven DA tasks. For Few-shot, three target samples are used. (SE) refers to the Shared encoder, while (DE) is the Disentangled encoders. (RL) specifies a model trained with the reconstruction loss.}}
\vspace{5pt}

\label{tab:da_results}
\resizebox{\textwidth}{!}{%
\begin{tabular}{l||c|c|c|c|c|c|c|c||c}
\multicolumn{1}{c||}{\textbf{Method}}
& \multicolumn{1}{c|}{\textbf{Type}} 
& \multicolumn{1}{c|}{\textbf{R} $\rightarrow$ \textbf{C}} 
& \multicolumn{1}{c|}{\textbf{R} $\rightarrow$ \textbf{P}} 
& \multicolumn{1}{c|}{\textbf{R} $\rightarrow$ \textbf{S}} 
& \multicolumn{1}{c|}{\textbf{P} $\rightarrow$ \textbf{C}} 
& \multicolumn{1}{c|}{\textbf{P} $\rightarrow$ \textbf{R}} 
& \multicolumn{1}{c|}{\textbf{C} $\rightarrow$ \textbf{S}} 
& \multicolumn{1}{c|}{\textbf{S} $\rightarrow$ \textbf{P}} 
& \multicolumn{1}{c}{\textbf{Average}} \\
\hline\hline
Source only & - & 56.59 $\pm$ 0.79 & 56.79 $\pm$ 0.50 & 46.25 $\pm$ 0.86 & 55.55 $\pm$ 0.83 & 66.20 $\pm$ 0.72 & 52.07 $\pm$ 1.01 & 44.81 $\pm$ 1.59 & 54.04 \\
TeachAugment\cite{Suzuki_2022_CVPR} & - & 53.84 $\pm$ 0.56                 
& 56.70 $\pm$ 0.59 & 46.70 $\pm$ 1.34 & 50.40 $\pm$ 1.27 & 58.64 $\pm$ 0.68 & 50.52 $\pm$ 0.09 & 44.89 $\pm$ 0.83 & 51.67 \\
\hline
ASM\cite{luo2020adversarial} & One-shot & 39.74 $\pm$ 0.56 & 46.39 $\pm$ 1.53 & 31.37 $\pm$ 5.51 & 4.31 $\pm$ 0.60 & 5.87 $\pm$ 2.33 & 37.12 $\pm$ 1.12 & 19.67 $\pm$ 2.99 & 26.35 \\
TOS-UDA \cite{carrazco2023targetdriven} & One-shot & \textbf{58.11} $\pm$ 0.38 & 58.57 $\pm$ 0.20 & \textbf{49.87} $\pm$ 0.97 & 54.24 $\pm$ 0.62 & 62.72 $\pm$ 0.32 & 52.88 $\pm$ 0.25 & 47.94 $\pm$ 1.12 & 54.90 \\
Our model (SE)
& One-shot & 49.89 $\pm$ 4.96 & 57.52 $\pm$ 7.05 & 39.07 $\pm$ 16.88 & 51.55 $\pm$ 3.27 & 58.08 $\pm$ 6.41 & 37.47 $\pm$ 19.76 & 42.09 $\pm$ 16.43 & 47.95 \\
Our model (DE+RL)
& One-shot & 56.74 $\pm$ 0.62 & 61.02 $\pm$ 1.31 & 47.03 $\pm$ 3.41 & 54.24 $\pm$ 2.66 & 69.06 $\pm$ 0.87 & 53.42 $\pm$ 0.76 & 52.95 $\pm$ 1.77 & 56.35 \\
\hline
Our model (SE) & Few-shot (3) & 57.06 $\pm$ 0.54 & 61.95 $\pm$ 0.57 & 49.18 $\pm$ 0.19 & 52.52 $\pm$ 3.01 & 66.79 $\pm$ 1.91 & 51.69 $\pm$ 1.99 & 50.08 $\pm$ 3.25 & 55.61 \\
Our model (DE+RL) & Few-shot (3) & 57.96 $\pm$ 0.81 & \textbf{62.43} $\pm$ 0.88 & 47.95 $\pm$ 1.09 & \textbf{56.70} $\pm$ 0.80 & \textbf{69.59} $\pm$ 0.41 & \textbf{55.37} $\pm$ 0.89 & \textbf{54.58} $\pm$ 0.97 & \textbf{57.80}
\end{tabular}
}
\end{table}
\subsection{Results}
In table \ref{tab:da_results}, the results for DomainNet are presented. we evaluate the accuracy of our method against four different competitors. \textbf{Source only}: a classifier trained only on source data using a cross entropy loss. \textbf{TeachAugment}\cite{Suzuki_2022_CVPR}: as explained in section \ref{sec:related_work}, this is a method for model generalization that learns fixed augmentations. \textbf{ASM}\cite{luo2020adversarial}. One of the few methods that tackles OS-UDA setting; it leverages style transfer to generate target samples. \textbf{TOS-UDA}\cite{carrazco2023targetdriven}: it implements a perceptual loss to guide the learning process to learn fixed transformations. Together with the aforementioned methods, we evaluate two variations of our model: the one using the shared encoder and the other with the disentangled encoders. 
The results shows that our method the disentangled encoders performs better than any of the other evaluated methods. Instead, our model with shared encoder obtained a lower accuracy than TOS-UDA. This could imply that one-shot is not enough for the shared encoder to transfer the style to the augmented sample. To expand our experiments, we also trained our method in the Few-shot setting (3 Target samples) to evaluate its accuracy if more samples were available. In this setting, our model with shared decoder increase its accuracy which proves that the shared encoder works only if more target samples are available. Our model with the disentangled encoders has an increase of (2.0\%) on its performance.
This indicates that our method could possibly be used for other DA scenarios where there are more available target samples, without the need to be modified.
\begin{table}[]
\centering
\caption{\small{Classification accuracy of our proposed method on VisDA. For Few-shot, three target samples are used. (SE) refers to the Shared encoder, while (DE) is the Disentangled encoders. (RL) specifies a model trained with the reconstruction loss.}}
\vspace{5pt}
\label{tab:visda_results}
\resizebox{\textwidth}{!}{%
\begin{tabular}{l|c|cccccccccccc|c}
\multicolumn{1}{c||}{\textbf{Method}}        
& \multicolumn{1}{c|}{\textbf{Type}} 
& \multicolumn{1}{c}{\textbf{Aeroplane}} 
& \multicolumn{1}{c}{\textbf{Bicycle}} 
& \multicolumn{1}{c}{\textbf{Bus}} 
& \multicolumn{1}{c}{\textbf{Car}} 
& \multicolumn{1}{c}{\textbf{Horse}} 
& \multicolumn{1}{c}{\textbf{Knife}} 
& \multicolumn{1}{c}{\textbf{Motorcycle}} 
& \multicolumn{1}{c}{\textbf{Person}} 
& \multicolumn{1}{c}{\textbf{Plant}} 
& \multicolumn{1}{c}{\textbf{Skateboard}} 
& \multicolumn{1}{c}{\textbf{Train}} 
& \multicolumn{1}{c|}{\textbf{Truck}} 
& \multicolumn{1}{c}{\textbf{Mean}} \\ \hline\hline
Source only & - & 68.86 & 3.24 & 46.05 & \textbf{97.61} & 30.48 & 8.08 & 50.69 & 5.90 & 72.14 & 16.97 & 62.21 & 14.84 & 39.76 \\
TeachAugm \cite{Suzuki_2022_CVPR} & - & 26.47 & 0.35 & 39.49 & 40.38 & 1.28 & 1.21 & 31.76 & 0.40 & 39.36 & 9.67 & 55.69 & 10.45 & 21.38 \\ \hline
ASM\cite{luo2020adversarial} & One-shot & 62.49 & 25.17 & \textbf{81.61} & 77.23 & 47.72 & 11.84 & 39.51 & 5.68 & \textbf{83.93} & 30.07 & 48.77 & \textbf{31.49} & 45.46 \\
TOS-UDA\cite{carrazco2023targetdriven} & One-shot & 15.05 & 0.01 & 13.96 & 17.31 & 2.47 & 20.38 & 0.53 & 1.34 & 11.46 & 7.15 & 20.51 & 5.32 & 9.63 \\
Ours (DE+RL) & One-shot & 59.90 & 12.77 & 71.99 & 91.46 & 48.44 & \textbf{23.70} & \textbf{59.88} & 11.56 & 76.38 & \textbf{40.22} & 63.19 & 24.26 & 48.64\\ \hline
ASM\cite{luo2020adversarial} & Few-shot (3) & \textbf{70.27} & \textbf{35.57} & 78.62 & 83.36 & 43.95 & 13.88 & 36.39 & 4.04 & 83.14 & 32.06 & 37.65 & 30.31 & 45.77\\
TOS-UDA\cite{carrazco2023targetdriven} & Few-shot (3) & 21.92 & 1.02 & 19.66 & 11.56 & 7.32 & 7.60 & 5.15 & 2.24 & 11.67 & 11.29 & 17.90 & 5.74 & 10.26 \\
Ours (DE+RL) & Few-shot (3) & 62.21 & 10.68 & 68.38 & 90.93 & \textbf{53.88} & 22.99 & 58.91 & \textbf{12.96} & 70.14 & 39.49 & \textbf{67.24} & 27.95 & \textbf{48.79}
\end{tabular}
}
\end{table}

In table \ref{tab:visda_results}, the results for VisDA are presented. we evaluated our proposed approach against the same methods that were used for DomainNet. To the best of our knowledge, it is the first time that VisDA is used as a benchmark for the OS-UDA setting. We generate the results presented for ASM and TOS-UDA following their respective training strategies. As before, we present the results using differnt number of target samples. For VisDA, we only tested the best performing configuration of our method, i.e. the disentangled encoders and reconstruction loss are used. Based on the results, we conclude that methods which use fixed transformations are not capable of closing the gap between synthetic and real domains. The aforementioned methods obtained lower accuracy than the source only method. Furthermore, our method with its disentangled encoders is able to obtained the highest mean accuracy among all evaluated methods. When using three target samples, our method obtains a high accuracy similar to the one-shot version. This could indicate that unlike the DomainNet experiments, the model would need to be modified for it to reach higher accuracy.
\subsection{Ablations}
\begin{table}[]
\centering
\caption{ \small{Comparative results for different ablations of our proposed approach are presented in the table. (SE) refers to Shared encoder, while (DE) represents the Disentangled encoders. (AvgP) indicates the use of average pooling by the Style Alignment module, and (RL) specifies a model trained with the reconstruction loss.}}
\vspace{5pt}
\label{tab:da_method_ablations}
\resizebox{\textwidth}{!}{%
\begin{tabular}{l||c|c|c|c|c|c|c||c}
\multicolumn{1}{c||}{\textbf{Model}}
& \multicolumn{1}{c|}{\textbf{R} $\rightarrow$ \textbf{C}} 
& \multicolumn{1}{c|}{\textbf{R} $\rightarrow$ \textbf{P}} 
& \multicolumn{1}{c|}{\textbf{R} $\rightarrow$ \textbf{S}} 
& \multicolumn{1}{c|}{\textbf{P} $\rightarrow$ \textbf{C}} 
& \multicolumn{1}{c|}{\textbf{P} $\rightarrow$ \textbf{R}} 
& \multicolumn{1}{c|}{\textbf{C} $\rightarrow$ \textbf{S}} 
& \multicolumn{1}{c||}{\textbf{S} $\rightarrow$ \textbf{P}} 
& \multicolumn{1}{c}{\textbf{Average}} \\
\hline\hline
SE & 56.53 $\pm$ 1.44 & 62.89 $\pm$ 1.03 & \textbf{50.33} $\pm$ 2.27 & 55.02 $\pm$ 1.96 & 64.32 $\pm$ 4.42 & 53.62 $\pm$ 0.73 & 50.33 $\pm$ 2.27 & 56.73 \\
SE+AvP & 54.16 $\pm$ 3.25 & 62.83 $\pm$ 0.40 & 48.64 $\pm$ 1.79 & 48.94 $\pm$ 5.24 & 66.21 $\pm$ 0.66 & 50.04 $\pm$ 4.69 & 52.91 $\pm$ 0.69 & 54.82 \\
DE & 56.12 $\pm$ 1.97 & \textbf{63.55} $\pm$ 0.61 & 49.87 $\pm$ 2.49 & 54.38 $\pm$ 2.00 & 66.53 $\pm$ 1.70 & 51.70 $\pm$ 1.41 & 54.34 $\pm$ 2.01 & 56.64\\
DE+AvP & 56.79 $\pm$ 0.84 & 62.50 $\pm$ 0.54 & 48.71 $\pm$ 0.66 & 56.18 $\pm$ 1.37 & 69.11 $\pm$ 0.61 & 54.50 $\pm$ 1.70 & \textbf{55.12} $\pm$ 0.61 & 57.56\\
DE+AvP+RL & \textbf{57.96} $\pm$ 0.81 & 62.43 $\pm$ 0.88 & 47.95 $\pm$ 1.09 & \textbf{56.70} $\pm$ 0.80 & \textbf{69.59} $\pm$ 0.41 & \textbf{55.37} $\pm$ 0.89 & 54.58 $\pm$ 0.97 & \textbf{57.80}
\end{tabular}
}
\end{table}
We conducted additional experiments to analyze various configurations of our model.  We compared our modified version of the perceptual loss, which incorporates an average pooling (AvgP) operation, with the original approach presented in \cite{johnson2016perceptual} that utilizes a Gram matrix (eq. \ref{eq:gram_matrix}). We tested two different architectures for AUM: a shared encoder with conditioning mixup (SE) and the disentangled encoders (DE). Additionally, for the second AUM architecture, we also examined its variation with the inclusion of a reconstruction loss (RL). In Table \ref{tab:da_method_ablations}, we present the results obtained from testing such ablations. \textbf{AUM architectures}. When comparing both architectures without any other variations, i.e. SE and DE, there is not a considerable difference in the average performance between them as the 0.09\% difference can be neglected. The same can be said about their evaluations on individual tasks as they differ only about 1.0\% in most of the task (6 out of 7). \textbf{Average pooling}. The use of average pooling by SAM shows opposite results based on the AUM architecture tested. we believe that the disentangled encoders is able to transfer more style information of the target sample than the shared encoder. therefore after applying average pooling, there is still enough information to measure the perceptual similarity for the style. \textbf{Reconstruction Loss}. Applying the reconstruction loss when training the AUM (DE) shows a slight increase in the average accuracy compare to the method which uses only the perceptual loss for training. However, this configuration of the model achieves the best results not only in average accuracy among all configurations but also obtains the best performances in four out of seven tasks.
\section{Conclusions}
\label{sec:conclusions}
In this paper, we introduced our novel approach called Learnable Data Augmentation for One-Shot Unsupervised Domain Adaptation (LearnAug-UDA). Our method addresses the difficult domain adaptation scenario known as OS-UDA, where the availability of only a single unlabeled target sample poses a challenge. To overcome this limitation, we developed an Augmentation Module that utilizes style transfer techniques to synthesize augmented samples. To guide the learning process of the AUM, we incorporated a Style Alignment Module, which enforces a perceptual similarity between the augmented samples and the target sample. We improved the perceptual loss by introducing an average pooling operation to smooth out the synthesized samples. Through these key components, our approach enables effective adaptation to the target domain. We tested our approach in the image classification task for two well-known Domain Adaptation benchmarks, DomainNet and Visda. We demonstrated the superior performance of our method compared to the selected baselines across a majority of the domain adaptation tasks. 
\bibliography{egbib}

\end{document}


\maketitle
	
\section{Introduction}
\label{sec:intro}
In this supplementary material, we expand the information presented for our proposed method LearnAug-UDA. In section \ref{sec:network_description}, we describe the network configuration for our Augmentation Module (AUM). In section  \ref{sec:qualitative_comparison}, we present a qualitative comparison of the augmented samples synthesized by our AUM and the baselines. In section \ref{sec:visda_results}, we expand the results presented for VisDA \cite{peng2017visda}.
\section{Encoder-Decoder description}
\label{sec:network_description}
Our proposed approach employs augmented samples that display perceptual similarities with the Target domain. These augmented samples are generated via an Augmentation module (AUM) which exploits style-transfer techniques to learn. We present two distinct versions of AUM, both of them based on an Encoder-Decoder architecture. The first version, the Shared Encoder (SE), consists of one encoder and one decoder architecture where the conditioning is done in the bottleneck via mixup \cite{zhang2017mixup}. The second version, the Disentangled Enconder (DE), consists of two encoders, one bottleneck module which mixes the embeddings from the encoders, and one decoder which synthesizes the augmented sample. In both versions, i.e. SE and DE, we make use of an encoder based on UNIT's encoder \cite{liu2017unsupervised}. For the decoder network, we based our architecture on UNIT's generator. Unlike UNIT, we change the deconvolutional layers of the decoder to an upsampling plus convolutional layer to minimize the Checkerboard artifacts on the augmented samples. Finally, the Bottleneck module is a convolutional block similar to the one used by the encoder. In Table \ref{tab:encoder_architecture}, we present the network architecture for the encoder, the decoder, and the bottleneck networks.
\begin{table}[ht]
\centering
\caption{\small{Network architecture for the Augmentation Module. The Shared Encoder and the Disentangled Encoders shared the same configurations for their respectives encoders.}}
\label{tab:encoder_architecture}
\tiny
\vspace{5pt}
\resizebox{0.8\textwidth}{!}{%
\begin{tabular}{ll}
\hline
\textbf{Layer} & \textbf{Encoder}                                                                \\ \hline
1              & Conv (channels=64, kernel size=7, stride=2), Leaky ReLU                         \\
2              & Conv (channels=128, kernel size=4, stride=2), Leaky ReLU                        \\
3              & Conv (channels=256, kernel size=4, stride=2), Leaky ReLU                        \\
4              & Residual block (channels=256, kernel size=3, stride=1)                          \\
5              & Residual block (channels=256, kernel size=3, stride=1)                          \\
6              & Residual block (channels=256, kernel size=3, stride=1)                          \\
7         & Residual block (channels=256, kernel size=3, stride=1)                          \\ \hline
\textbf{Layer} & \textbf{Decoder}                                                                \\ \hline
1              & Residual block (channels=256, kernel size=3, stride=1)                          \\
2              & Residual block (channels=256, kernel size=3, stride=1)                          \\
3              & Residual block (channels=256, kernel size=3, stride=1)                          \\
4              & Residual block (channels=256, kernel size=3, stride=1)                          \\
5              & Upsampling (Bilinear), Conv (channels=128, kernel size=3, stride=1), Leaky ReLU \\
6              & Upsampling (Bilinear), Conv (channels=128, kernel size=3, stride=1), Leaky ReLU \\
7              & Conv (channels=3, kernel size=3, stride=1), Sigmoid                             \\ \hline
\textbf{Layer} & \textbf{Bottleneck}                                                             \\ \hline
1              & Conv (channels=256, kernel size=7, stride=1), ReLU                              \\
               &                                                                                
\end{tabular}
}
\end{table}
\section{Qualitative comparisons}
\label{sec:qualitative_comparison}
In this section, we present a comparison between the diverse augmented samples generated by our method and the baselines, i.e. ASM\cite{luo2020adversarial}, TeachAugment\cite{Suzuki_2022_CVPR}, and TOS-UDA\cite{carrazco2023targetdriven}. To facilitate a comprehensive comparison, all the methods were trained using the same target samples except for TeachAugment, i.e. TeachAugment does not requires target data.
\subsection{DomainNet (1 Target)}
\label{sub:qualitative_DNMethods}
\begin{figure}[ht]
\centering
\includegraphics[width=0.7\textwidth]{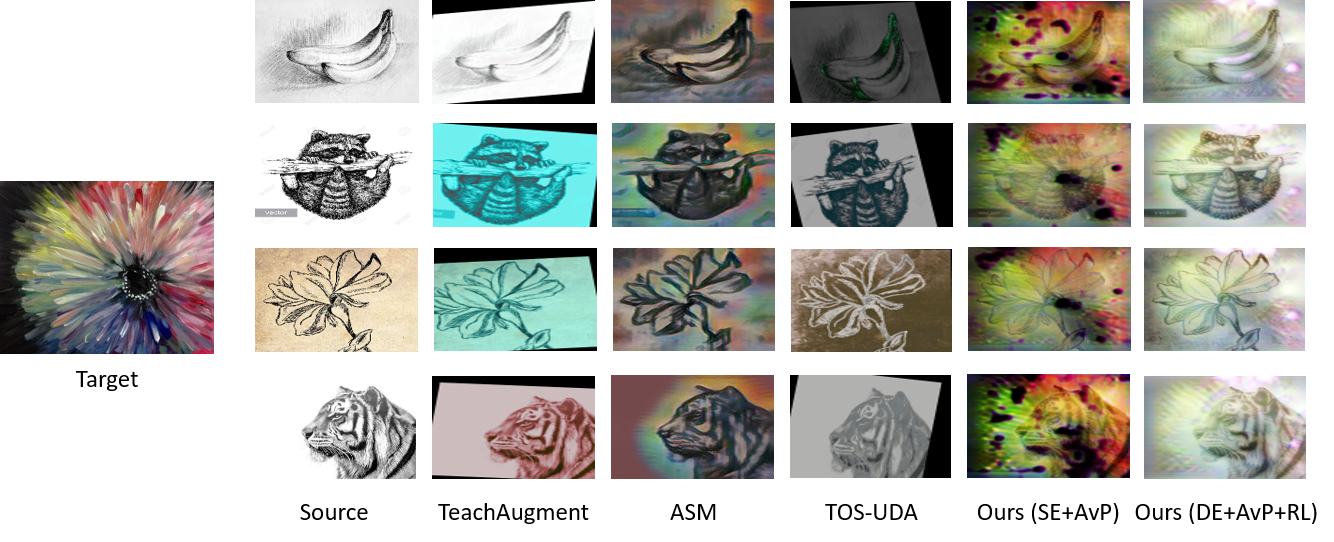}
\caption{\small{Qualitative  comparison between our proposed approach and the selected baselines. (SE) refers to Shared encoder, while (DE) represents the Disentangled encoders. (AvgP) indicates the use of average pooling by the Style Alignment module, and (RL) specifies a model trained with the reconstruction loss.}}
\label{fig:1T_DNMethods}
\end{figure}
In Figure \ref{fig:1T_DNMethods}, we illustrate a set of augmented samples synthesized by our proposed approach and the selected baselines. The augmeted samples are synthesized for the DA task of Sketch to Painting of DomainNet. By choosing this DA task, we are able to display the range of the possible augmentations that each methods is capable of. The augmented samples of TOS-UDA and TeachAugment are not capable of properly represent the color spectrum of the target image as they only work with fixed transformations. For ASM, its augmented samples display a perceptual similarity closer to target. However, ASM utilizes a pretrained module (RAIN) on WikiArts which results in an advantage when evaluating this specific DA task (Sketch to Painting). ASM may not have the same results for other domains. Furthermore, our augmented samples are generated by the Augmentation module which does not require pretraining to synthesize augmented samples with high perceptual similarity to the target.
\subsection{Method ablations}
\label{sub:qualitative_ablation}
\begin{figure}[h]
\centering
\includegraphics[width=0.7\textwidth]{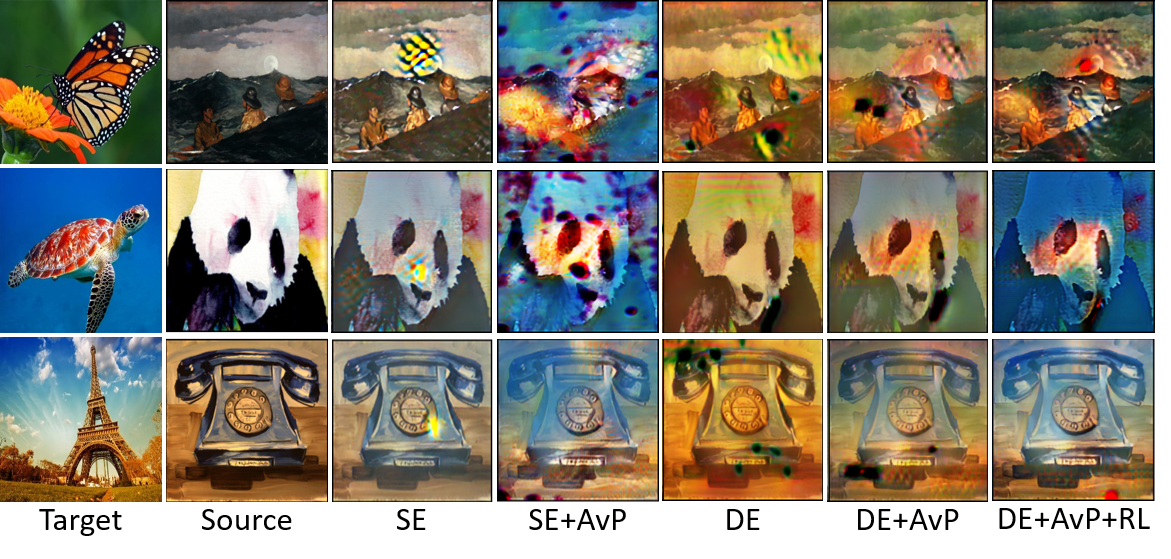}
\caption{\small{Qualitative  comparison between different ablations of our proposed approach.}}
\label{fig:qualitative_ablation}
\end{figure}

In Figure \ref{fig:qualitative_ablation}, we present different augmented samples that were synthesized using different ablations of the Augmentation module (AUM). For this comparison, we trained our proposed method using three target samples (see Fig. \ref{fig:qualitative_ablation} Target). These augmented samples are synthesized for the DA task of Painting to Real of DomainNet. In Table \ref{tab:P2R_accuracies}, we present the reported accuracies for this specific DA task to allow a better comparison of the augmented samples. Now, the presented images clearly demonstrate that applying the average pooling operation helps to smooth out hard details that are transferred from the target samples. Furthermore, the Disentangled encoders (DE) are capable of synthesizing images with less artifacts than the Shared encoder (SE), i.e. the augmented samples are less noisy therefore it obtains a better performance. Finally, introducing the reconstruction loss (RL) into the process allows the AUM to disentangle better content and style. Thus, the style encoder is capable of transferring better the characteristics of the Target domain.
\begin{table}[h]
\centering
\footnotesize
\caption{\small{Reported accuracies for the DA task of Painting to Real for DomainNet. (SE) refers to the Shared encoder, while (DE) is the Disentangled encoders. (AvgP) indicates the use of average pooling by the Style Alignment module, and (RL) specifies a model trained with the reconstruction loss.}}
\label{tab:P2R_accuracies}
\vspace{5pt}
\resizebox{0.8\textwidth}{!}{%
\begin{tabular}{c|ccccc}
         & \textbf{SE} & \textbf{SE+AvP} & \textbf{DE} & \textbf{DE+AvP} & \textbf{DE+AvP+RL} \\ \hline
Accurary & 64.32 $\pm$ 4.42 & 66.21 $\pm$ 0.66 & 66.53 $\pm$ 1.70 & 69.11 $\pm$ 0.61 & 69.59 $\pm$ 0.41             
\end{tabular}
}
\end{table}
\section{VisDA results}
\label{sec:visda_results}
In Table \ref{tab:visda_results}, we present the results for VisDA \cite{peng2017visda}. The results are obtained after performing five experiments for each of the methods. The target for each experiment was selected randomly. We present mean accuracy over all the class and their corresponding standard deviations. The results indicates that VisDA is a more challenging DA benchmark. The presence of large standard deviation values, particularly in certain classes, suggests that the quality of the selected target has a profound effect on the synthesized samples. However, upon observing the mean accuracy and its standard deviation, we can conclude that the proposed method consistently performs well.
\begin{sidewaystable}[ht]
\tiny
\centering
\caption{\small{Classification accuracy of our proposed method on VisDA. For Few-shot, three target samples are used. (SE) refers to the Shared encoder, while (DE) is the Disentangled encoders. (RL) specifies a model trained with the reconstruction loss.}}
\vspace{5pt}
\label{tab:visda_results}
\begin{tabular}{l|c|cccccccccccc|c}
\textbf{Method}           
& \multicolumn{1}{l|}{\textbf{\#.T.}} 
& \multicolumn{1}{l}{\textbf{Aeroplane}} 
& \multicolumn{1}{l}{\textbf{Bicycle}} 
& \multicolumn{1}{l}{\textbf{Bus}} 
& \multicolumn{1}{l}{\textbf{Car}} 
& \multicolumn{1}{l}{\textbf{Horse}} 
& \multicolumn{1}{l}{\textbf{Knife}} 
& \multicolumn{1}{l}{\textbf{Motorcycle}} 
& \multicolumn{1}{l}{\textbf{Person}} 
& \multicolumn{1}{l}{\textbf{Plant}} 
& \multicolumn{1}{l}{\textbf{Skateboard}} 
& \multicolumn{1}{l}{\textbf{Train}} 
& \multicolumn{1}{l|}{\textbf{Truck}} 
& \multicolumn{1}{l}{\textbf{Mean}} \\ \hline\hline
Source only & - & \textbf{68.86} $\pm$ 8.41 & 3.24 $\pm$ 0.96 & 46.05 $\pm$ 8.73 & \textbf{97.61} $\pm$ 1.15 & 30.48 $\pm$ 10.91 & 8.08 $\pm$ 3.44 & 50.69 $\pm$ 8.98 & 5.90 $\pm$ 3.66 & 72.14 $\pm$ 18.22 & 16.97 $\pm$ 4.42 & 62.21 $\pm$ 11.79 & 14.84 $\pm$ 5.14 & 39.76 $\pm$ 5.38 \\
TeachAugm \cite{Suzuki_2022_CVPR} & - & 26.47 $\pm$ 3.18 & 0.35 $\pm$ 0.24 & 39.49 $\pm$ 11.79 & 40.38 $\pm$ 8.59 & 1.28 $\pm$ 0.59 & 1.21 $\pm$ 0.62 & 31.76 $\pm$ 8.75 & 0.40 $\pm$ 0.27 & 39.36 $\pm$ 11.45 & 9.67 $\pm$ 1.72 & 55.69 $\pm$ 17.22 & 10.45 $\pm$ 7.94 & 21.38 $\pm$ 1.49 \\ \hline
ASM\cite{luo2020adversarial} & 1 & 62.49 $\pm$ 9.51 & \textbf{25.17} $\pm$ 7.22 & \textbf{81.61} $\pm$ 4.38 & 77.23 $\pm$ 5.20 & 47.72 $\pm$ 10.28 & 11.84 $\pm$ 3.74 & 39.51 $\pm$ 12.10 & 5.68 $\pm$ 1.36 & \textbf{83.93} $\pm$ 7.87 & 30.07 $\pm$ 7.08 & 48.77 $\pm$ 11.12 & \textbf{31.49} $\pm$ 7.37 & 45.46 $\pm$ 1.24 \\
TOS-UDA\cite{carrazco2023targetdriven} & 1 & 15.05 $\pm$ 12.49 & 0.01 $\pm$ 0.02 & 13.96 $\pm$ 15.88 & 17.31 $\pm$ 17.19 & 2.47 $\pm$ 3.80 & 20.38 $\pm$ 34.03 & 0.53 $\pm$ 0.29 & 1.34 $\pm$ 1.63 & 11.46 $\pm$ 15.11 & 7.15 $\pm$ 7.84 & 20.51 $\pm$ 17.48 & 5.32 $\pm$ 6.46 & 9.63 $\pm$ 6.46 \\
Ours (DE+RL) & 1 & 59.90 $\pm$ 6.54 & 12.77 $\pm$ 3.61 & 71.99 $\pm$ 10.45 & 91.46 $\pm$ 3.02 & 48.44 $\pm$ 5.83 & \textbf{23.70} $\pm$ 7.29 & \textbf{59.88} $\pm$ 5.69 & 11.56 $\pm$ 4.55 & 76.38 $\pm$ 5.99 & \textbf{40.22} $\pm$ 2.00 & 63.19 $\pm$ 8.82 & 24.26 $\pm$ 4.87 & 48.64 $\pm$ 2.56\\ \hline
TOS-UDA\cite{carrazco2023targetdriven} & 3 & 21.92 $\pm$ 18.24 & 1.02 $\pm$ 1.76 & 19.66 $\pm$ 11.76 & 11.56 $\pm$ 20.25 & 7.32 $\pm$ 11.27 & 7.60 $\pm$ 14.49 & 5.15 $\pm$ 7.04 & 2.24 $\pm$ 4.14 & 11.67 $\pm$ 2.66 & 11.29 $\pm$ 4.05 & 17.90 $\pm$ 21.06 & 5.74 $\pm$ 5.74 & 10.26 $\pm$ 1.61 \\
Ours (DE+RL) & 3 & 62.21 $\pm$ 9.09 & 10.68 $\pm$ 3.02 & 68.38 $\pm$ 4.20 & 90.93 $\pm$ 3.06 & \textbf{53.88} $\pm$ 3.85 & 22.99 $\pm$ 2.79 & 58.91 $\pm$ 7.16 & \textbf{12.66} $\pm$ 2.85 & 70.14 $\pm$ 4.15 & 39.49 $\pm$ 4.28 & \textbf{67.24} $\pm$ 2.46 & 27.95 $\pm$ 3.48 & \textbf{48.79} $\pm$ 0.50
\end{tabular}
\end{sidewaystable}
\clearpage
\bibliography{egbib}